\documentclass[sigconf]{acmart}
\usepackage{graphicx}
\usepackage{tcolorbox}
\tcbuselibrary{breakable}

\AtBeginDocument{%
  }

\acmSubmissionID{1051}

\begin{document}

\title{Multi-modal Synthetic Data Training and Model Collapse: Insights from VLMs and Diffusion Models}

\settopmatter{printacmref=false}
\setcopyright{none}
\renewcommand\footnotetextcopyrightpermission[1]{}
\pagestyle{plain}

\author{Zizhao Hu}
\orcid{YourORCID}
\affiliation{%
  \institution{University of Southern California}
  \city{Los Angeles}
  \state{California}
  \country{USA}
}
\email{zizhaoh@usc.edu}
\author{Mohammad Rostami}
\orcid{TheirORCID}
\affiliation{%
  \institution{University of Southern California}
  \city{Los Angeles}
  \state{California}
  \country{USA}
}
\email{rostamim@usc.edu}
\author{Jesse Thomason}
\orcid{AnotherORCID}
\affiliation{%
  \institution{University of Southern California}
  \city{Los Angeles}
  \state{California}
  \country{USA}
}
\email{jessetho@usc.edu}



\renewcommand{\shortauthors}{Hu et al.}

\begin{abstract}
Recent research has highlighted the risk of generative model collapse, where performance progressively degrades when continually trained on self-generated data. However, existing exploration on model collapse is limited to single, unimodal models, limiting our understanding in more realistic scenarios, such as diverse multi-modal AI agents interacting autonomously through synthetic data and continually evolving. We expand the synthetic data training and model collapse study to multi-modal vision-language generative systems, such as vision-language models (VLMs) and text-to-image diffusion models, as well as recursive generate-train loops with multiple models. We find that model collapse, previously observed in single-modality generative models, exhibits distinct characteristics in the multi-modal context, such as improved vision-language alignment and increased variance in VLM image-captioning task. Additionally, we find that general approaches such as increased decoding budgets, greater model diversity, and relabeling with frozen models can effectively mitigate model collapse. Our findings provide initial insights and practical guidelines for reducing the risk of model collapse in self-improving multi-agent AI systems and curating robust multi-modal synthetic datasets. 
\end{abstract}

\maketitle

\begin{figure*}[ht]
    \centering
    \includegraphics[width=\linewidth]{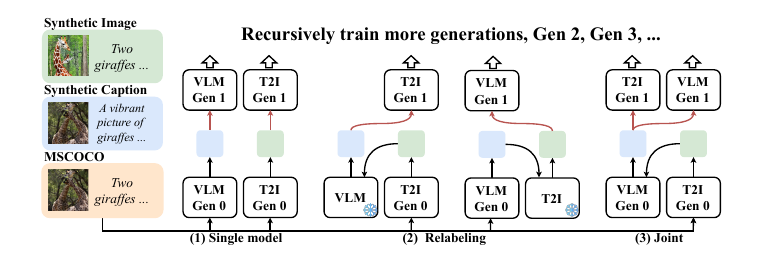}
    \caption{Recursive finetuning configurations for investigating model collapse: (1) Single model recursive finetuning (``T2I'' stands for text-to-image diffusion model),  (2) Finetuning with VLM or diffusion relabeling, and (3) Joint recursive finetuning. (1) allows independent study of the effect of recursive training on VLMs and text-to-image diffusion models. (2) enables information integration from another generative paradigm, which simulates a multi-agent environment where a frozen grounding model is available and used to label synthetic data. (3) simulates a multi-agent environment where all models have full freedom to update their weights using available synthetic data. }
    \Description{Diagram showing three different experimental setups for recursive finetuning of vision-language models.}
    \label{fig:settings}
\end{figure*}

\section{Introduction}
The key to training large-scale generative AI models is the availability of large-scale training datasets.
As models grow in size and capability, the amount of available human-authored data for training becomes a significant bottleneck. This bottleneck has prompted the exploration of synthetic data for training AI models. In the realm of LLMs, synthetic data has been used for pretraining~\cite{phi1,phi1.5,phi3}, continued pre-training ~ \cite{yang2024syntheticcontinuedpretraining}, supervised finetuning for general tasks~ \cite{li2024syntheticdataalmostscratch}, or specialized tasks such as math
~\cite{yu2024metamathbootstrapmathematicalquestions, Luo2023WizardMathEM,gao2023gllavasolvinggeometricproblem, Trinh2024SolvingOG} and coding~\cite{magicoder, wizardcoder}, and post-training RL ~\cite{deepseekr1, llama,qwen,tulu}. Other than LLMs, synthetic data has also been used to train VLMs~\cite{llava}. In continual learning, synthetic data has also been used to reduce catastrophic forgetting~\cite{gr}. 

In model training, synthetic data offers advantages in terms of quantity, controllability, and privacy. However, because generative models are often used to generate synthetic data for training generative models, model collapse through recursive training on self-generated data becomes a risk~\cite{shumailov2023curse, alemohammad2023self}. We directly address several limitations of the existing literature studying such generative model collapse.
In particular, existing works:

\begin{enumerate}
    \item focus on characterizing model collapse in single-modality tasks (e.g., language generation or unconditional image generation), providing limited insight into how collapse manifests in multi-modal generative systems;

    \item focus on a single model's self-consuming training loop, without providing insights on more realistic scenarios where multiple models, from multiple generative paradigms, are within such a loop; and
    
    \item offer solutions to model collapse primarily dependent on mixing synthetic data with human-authored data, without exploring how to improve the robustness of synthetic data. 
\end{enumerate}

These limitations are particularly concerning given that a growing proportion of labeled web content is generated by two classes of multi-modal models: text-to-image diffusion models~ \cite{rombach2022high, ramesh2021zero, ramesh2022hierarchical, BetkerImprovingIG} and VLMs ~ \cite{llava, blip2, gpt4o,vila,qwenvl,internvl25}. In this study, we address these limitations by expanding the investigation of synthetic data training and model collapse to multi-modal, multi-agent scenarios, answering:

\begin{enumerate}
\item How does recursive synthetic data training impact multi-modal generative systems?
\item What properties characterize multi-modal model collapse?
\item Do model interactions make collapse better or worse? 
\item How can we improve generative synthetic data to mitigate model collapse?
\end{enumerate}

Through extensive experiments on VLMs and diffusion models, we make several key discoveries:

\begin{enumerate}
\item Balancing synthetic data bias-variance trade-off should be the focus for reducing model collapse rather than increasing variance only, since variance does not always collapse for certain tasks, such as VLM image captioning;
\item increased decoding budgets generate more robust synthetic datasets, which can mitigate model collapse despite no significant change in data quality;
\item commonly used statistical quality measures correlate with the robustness of synthetic data against model collapse; and
\item adding more models to the recursive training loop can make the model collapse worse, unless the added models are frozen and pretrained on human-authored data.
\end{enumerate}

Our analysis centers around recursive synthetic data training under two paradigms: image-captioning VLMs and text-to-image diffusion models. Our work provides insight into the generate-train self-consuming loop in a multi-modal and multi-agent environment. 
 
\section{Related work}
In this section, we briefly cover some representative existing works studying model training with synthetic data and model collapse, and provide background on multi-modal generative models.\\\\
\textbf{Model training using generative synthetic data.} Generative models have been used for synthetic data curation for various downstream NLP tasks~\cite{yu2024metamathbootstrapmathematicalquestions, Luo2023WizardMathEM,gao2023gllavasolvinggeometricproblem, Trinh2024SolvingOG,magicoder, wizardcoder} and vision tasks  \cite{shin2018medical,coyner2022synthetic,daum2024differentially}.\\\\
\textbf{Model collapse.}
Model collapse is a concept first coined by Shumailov et al.~ \cite{shumailov2023curse}. It is a phenomenon in which models recursively trained on self-generated data experience performance degradation. It is hypothesized that model collapse happens because synthetic data lacks the diversity and nuances of real-world data. This issue has been identified in both language models  \cite{modelcollapse,shumailov2023curse,seddik2024badtrainingsyntheticdata} and image-generation models~ \cite{alemohammad2023self}. These works characterize model collapse by reduced perplexity and increased repetition in outputs generated by LLMs and diversity and quality degradation in output images of diffusion models. Addressing model collapse is critical, considering that using AI-generated synthetic data for training large AI models is becoming unavoidable as we exhaust human-authored data.\\\\
\textbf{Text-to-image diffusion models and VLMs.}
Text-to-image models like DALL-E  \cite{ramesh2021zero, ramesh2022hierarchical, BetkerImprovingIG}, Stable Diffusion  \cite{rombach2022high}, Imagen~\cite{imagen3}, and Midjourney have revolutionized image generation because they enable users to generate high-quality images directly from textual descriptions. VLMs such as CLIP  \cite{radford2021learning} and BLIP  \cite{blip, blip2} have advanced vision-language understanding. The broader class of VLMs or multi-modal LLMs can handle more diverse tasks, including VQA~\cite{vqa}, semantic segmentation~\cite{ssam}, and even autoregressive image generation~\cite {llamagen}.
However, we concentrate our study of VLMs' model collapse on the image-captioning task.



\section{Experiment settings}
Here, we detail the specific setups for the model collapse experiment and the model diversity experiment in separate subsections. Then, we describe the evaluation methods employed in our analysis. Finally, we explain the training settings and dataset used.
\subsection{Model collapse}
\label{subsec:model_collapse_analysis}
We consider three settings in which to analyze model collapse (see Figure \ref{fig:settings}). In setting 1, we study the characteristics of model collapse in a single diffusion model or VLM. In settings 2 and 3, we study the interaction between VLMs and diffusion models. Setting 2 adds a frozen VLM or diffusion model to relabel the diffusion or VLM synthetic data in the recursive loop. Setting 3 makes the frozen models in setting 2 also trainable, simulating a joint generating-training environment. For detailed data processing, hyperparameter setting, and model specifications, see Appendix~\ref{appendix:experimental_details}. \\\\
\textbf{Setting 1: Single model recursive fine-tuning.} 
 For the diffusion model, we use Stable Diffusion 1.5~\cite{rombach2022high}. Images are generated from a fixed subset of MSCOCO training captions. Newly generated images are paired with the original captions to fine-tune the model in the next generation. For the VLM, we use BLIP-2~\cite{blip2}, and the captions are generated from a fixed subset of MSCOCO training, guided by a constant prefix prompt. Newly generated captions are paired with the original images to fine-tune the model in the next generation. The results and analysis for this setting are presented in Sections~\ref{subsec:diffusion_collapse}--\ref{subsec:decoding_budgets}, where we investigate model collapse characteristics in diffusion models and VLMs, variance behaviors, and the effects of hyperparameters and decoding budgets.\\\\
\textbf{Setting 2: Two-model recursive fine-tuning with frozen relabeling.}
\label{setting:relabeling}
This setting simulates a scenario where a model of a different paradigm trained on human-authored data serves as a labeler to relabel the synthetic data. The two models that we use are the same as setting 1, but we enable interaction between the VLM and the Diffusion model. Specifically, before passing the generated synthetic data to the next generation of training, we use a frozen VLM or Diffusion model to relabel the generated images or captions. Consequently, the conditioning modalities are updated for every generation. The effectiveness of this ``relabeling'' approach in mitigating model collapse is analyzed in Section~\ref{subsec:relabeling}.\\\\
\textbf{Setting 3: Two-model joint recursive fine-tuning.}
\label{setting:joint} This setting simulates a scenario where a VLM and a diffusion model both dynamically evolve on synthetic data. This setting is similar to setting 2, except the grounding model is first fine-tuned on the synthetic data pairs before providing relabeling for the next generation. The results for this joint fine-tuning approach and its impact on model collapse are presented in Section~\ref{subsec:joint_training}.

\subsection{Model diversity}
\label{subsec:diversity_experiments}
To study how model diversity affects model collapse, we set up two different types of diversity settings: 1. hyperparameter diversity and 2. architectural diversity, defined as follows: \\\\
\textbf{Hyperparameter diversity.} Compared with the baseline settings, we use different hyperparameters to generate the same amount of data for synthetic data training. For fair comparison, we choose the hyperparameters that do not impact decoding cost. For image generation, we use 5 different classifier-free guidance (CFG) levels (1,3,7,10,20) of Stable Diffusion 1.5, to generate 1000 images for each generation, with 200 images each. For image-captioning, we do the same with 5 different temperature values (0.6, 0.7, 0.8, 0.9, 1). 
\\\\
\textbf{Architectural diversity.} We use additional diffusion models (Stable Diffusion 1.4 and Flux.1) and additional VLMs (BLIP, MiniCPM-Llama3-V 2.5) in our experiments. To simplify the process, the additional models are frozen and only used for generating synthetic data, which is mixed with the synthetic data generated during the BLIP-2 and Stable Diffusion 1.5 recursive finetuning process.  

\begin{figure*}[ht]
\centering
\includegraphics[width=\linewidth]{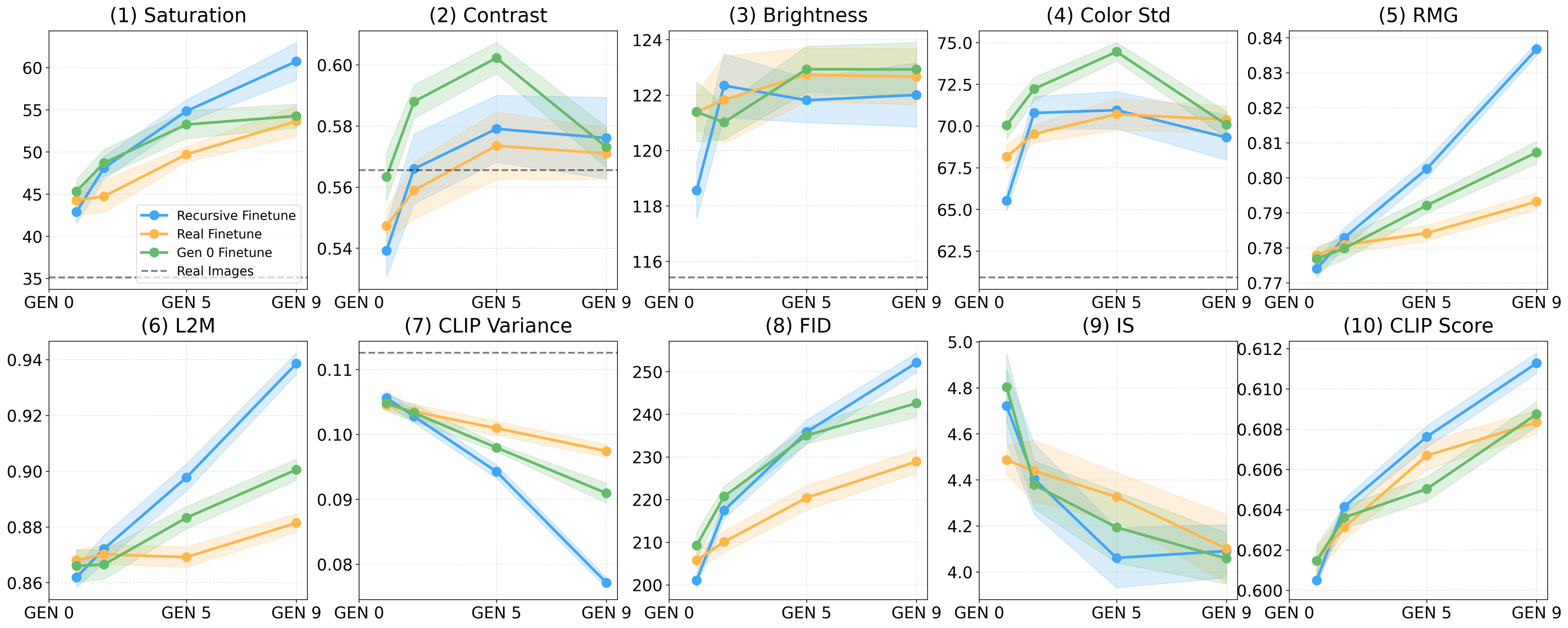}
\caption{Recursive finetuning shifts generated image properties. Mean (solid lines) and standard deviation (shaded area) are calculated from 5 groups of 200 evaluation samples. Two baselines are shown: ``Gen 0 Finetune'' stands for finetuning only on the synthetic data generated by the generation 0 (pretrained) model. ``Real Finetune'' stands for finetuning on the human-authored data. The most prominent effect of recursive finetuning compared to the other two baselines is the shifted saturation (1), reduced CLIP embedding variance (7), increased modality gaps (5, 6), and increased CLIP Score (10). }
\Description{Graph showing various image metrics across generations of recursive finetuning, highlighting changes in saturation, embedding variance, and modality gaps.}
\label{fig:image_metrics}
\end{figure*}

\subsection{Evaluation settings}
We use a comprehensive set of evaluation metrics, aiming to pinpoint the key characteristics of model collapse.\\\\
\textbf{Inherent properties.}
We use saturation, contrast, brightness, color standard deviation for images, and average length, and vocabulary size for captions. For definitions please refer to Appendix Section~\ref{appendix:metrics}. \\\\
\textbf{Repensentational variance.}
To quantify the variance of a modality, we calculate its CLIP embedding variance. The shared CLIP embedding space allows direct comparison of the image and text. \\\\
\textbf{Quality.}
We use FID~\cite{heusel2017gans} and IS  \cite{zhang2018unreasonable} for images,
and BLEU~\cite{papineni2002bleu}, METEOR  ~\cite{banerjee2005meteor}, and BERTScore  ~\cite{zhang2019bertscore} for captions.
For definitions please refer to Appendix Section~\ref{appendix:metrics}. FID and BLEU are also used to quantify model collapse intensity.\\\\
\textbf{Vision-language alignment.}
We use CLIP Score~\cite{radford2021learning}  to measure the alignment between captions and images. Following the modality gap ~\cite{liang2022mind} study in contrastive VLMs, we also included L2 modality gap (L2M)~\cite{liang2022mind} and relative modality gap (RMG)~\cite{Schrodi2024TwoEO}. These metrics allow an understanding of the embedding distribution shift. Since the modality gap has been found to correlate with information imbalance between modalities~\cite{Schrodi2024TwoEO}, we use it to measure information balance change between modalities. \\\\
\textbf{Gender biases.}
When diffusion models are given a gender-neutral prompt about an occupation, and asked to generate multiple images of a person in this occupation, the genders of the generated persons might be biased towards male or female. To study how this bias changes during recursive finetuning, we use a gender classifier (described in Appendix Section~\ref{appendix:metrics}) to classify the gender of the person in the recursively generated images, given 100 gender neutral prompts for 10 different occupations each. The classifier outputs male, female. The ratio of female and male predictions is then calculated, and the metric is denoted as $n-B$, where $n$ is a ratio ranging from $0.5$ to $1$, with $B$ indicating the direction of bias: $M$ signifying a bias towards male and $F$ towards female.

\subsection{Training settings} 
We employ full finetuning to update all parameters for all experiments. While efficient low-resource finetuning approaches such as LoRA~ \cite{hu2022lora} are viable solutions for updating large models, we opt for full model finetuning to study model collapse. Full model finetuning allows our findings to be extendable to continued pretraining scenarios. In our model collapse investigation, we consider a ``\textbf{Recursive Finetuning}'' setting, where the model generates data, fine-tunes on the generated data, and continues this loop for several generations. In addition, we have two baseline settings: ``\textbf{Real Finetuning}'', where the synthetic data is replaced with the human-authored data, and ``\textbf{Gen 0 Finetuning }'', where the recursively generated synthetic data is replaced with the synthetic data generated by the ``generation 0'' model (i.e., the pretrained model). 
\subsection{Dataset} 
We use an MSCOCO subset of 1000 samples for creating initial conditions for generation and setting up the fine-tuning baselines. This also enables meaningful evaluation on the MSCOCO evaluation set. For the subset configurations, please see Section~\ref{appendix:data_preprocessing}
\section{Analysis}
We first study the characteristics of model collapse in text-to-image diffusion models (Section~\ref{subsec:diffusion_collapse}) and VLMs (Section~\ref{subsec:vlm_collapse}). We then challenge the commonly believed universal variance reduction in model collapse by providing evidence that it does not happen in VLM image-captioning tasks (Section~\ref{subsec:variance}). Next, we study what aspects of synthetic data can slow down model collapse (Section~\ref{subsec:bias_variance}). We also show that decoding budgets correlate with less model collapse, shown as better FID at recursive generation 10 (Section~\ref{subsec:decoding_budgets}). Finally, we provide several insights on how to reduce model collapse with synthetic data using diverse model interactions (Sections~\ref{subsec:model_diversity}--\ref{subsec:joint_training}).

\subsection{Characteristics of model collapse in text-to-image diffusion models}
\label{subsec:diffusion_collapse}

\begin{figure}[ht]
    \centering
    \includegraphics[width=\linewidth]{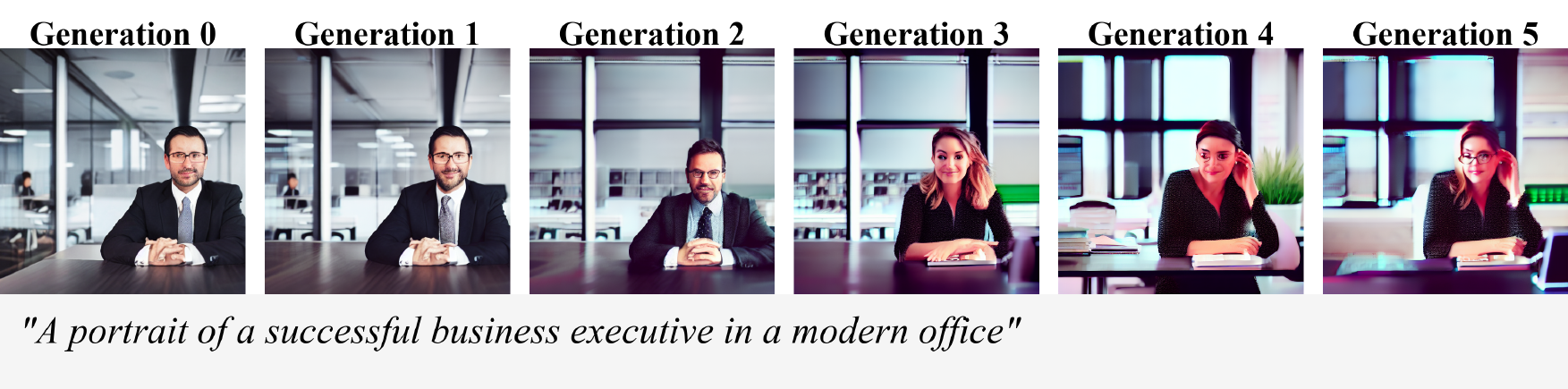}
    \caption{Gender bias shift: In this example, we observe a shift towards females in tandem with saturation upshift.}
    \Description{Graph demonstrating the relationship between image saturation and gender bias in generated images.}
    \label{fig:gender-bias}
\end{figure}

\begin{figure}[ht]
    \centering
    \includegraphics[width=\linewidth]{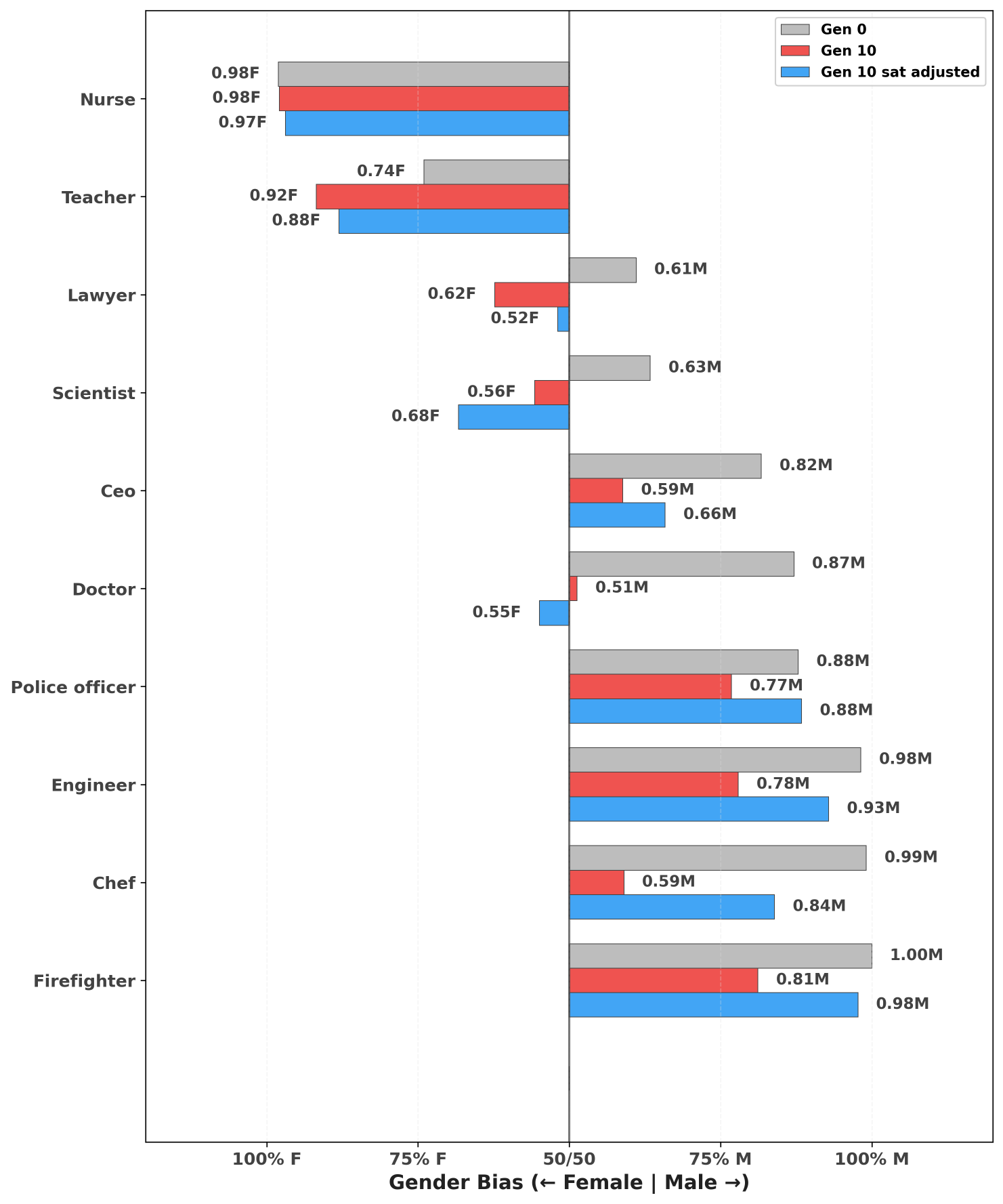}
    \caption{Gender bias shift: we see a shift to the female side after 10 generations of recursive finetuning on the diffusion model. After manually downshifting the saturation after each generation, we see a reduced gender bias shift, indicating that saturation drives the gender bias shift in recursive finetuning of the Stalbe Diffusion 1.4 model.}
    \Description{Graph showing how gender representation in generated images changes over generations of recursive finetuning.}
    \label{fig:gender-bias-bar}
\end{figure}

\textbf{Loss of variance but improved alignment.}
  We observe that recursive finetuning has a similar trend as ``Gen 0 Finetuning'', but with more significant shifts in terms of saturation, RMG, L2M, CLIP variance, FID, and CLIP Score (Figure~\ref{fig:image_metrics}). Among these metric shifts, the most prominent is that the image representational variance collapses faster in recursive finetuning than the other two baselines, as shown by a 4~times faster CLIP variance drop rate compared to MSCOCO data finetuning. The constantly improved caption-image alignment is driven by this reduced variance. This effect removes objects and high-frequency details that are not present in the captions.
  These effects can be seen in Figure  \ref{fig:cats} as the details in the backgrounds vanish. 
  \\\\
\textbf{Increased modality gaps.}
Both RMG (Figure \ref{fig:image_metrics} subfigure 5) and L2M (subfigure 6) show an increasing trend. However, such increases in modality gaps do not cause a drop in CLIP Score, which is a pair-wise measure. These combined suggest that the generated image distribution centroid shifts away from the centroid of the caption, but pair-wisely are more aligned with the captions.
\\\\
\textbf{Saturation-shift driven gender biases.}
We observe an up-shift of image saturation in all finetuning settings, where a more drastic change for recursive finetuning is observed. This up-shift trend is established as early as generation 1. We also observed a gender shift in the generated images across generations (Figure \ref{fig:gender-bias}), although the initial finetuning captions do not contain gendered words or information. To quantify the gender bias shift, we selected 10 occupations as prompts for the diffusion model to generate 100 images for each. Then, we used a pretrained gender classifier to predict the genders for each image. Figure \ref{fig:gender-bias-bar} shows that compared to the model at generation 0, there is a general shift towards females, even in cases where female representation was already heavily biased. This observation indicates that such a shift is not driven by the distributional properties of the finetuning set, but by some unidirectional drifting properties in the finetuning process. We hypothesize that the up-shift of saturation is more related to female and female-confounding concepts. To test this hypothesis, we manually tuned down the saturation after each recursive finetuning generation for the synthetic images (Figure \ref{fig:gender-bias-bar} ``Gen 10 sat adjusted''). We observe that for most occupations, the gender bias was reversed towards the distribution at generation 0, confirming our hypothesis.

\subsection{Characteristics of model collapse in VLMs}
\label{subsec:vlm_collapse}
\begin{figure*}[ht]
    \centering
    \includegraphics[width=\linewidth]{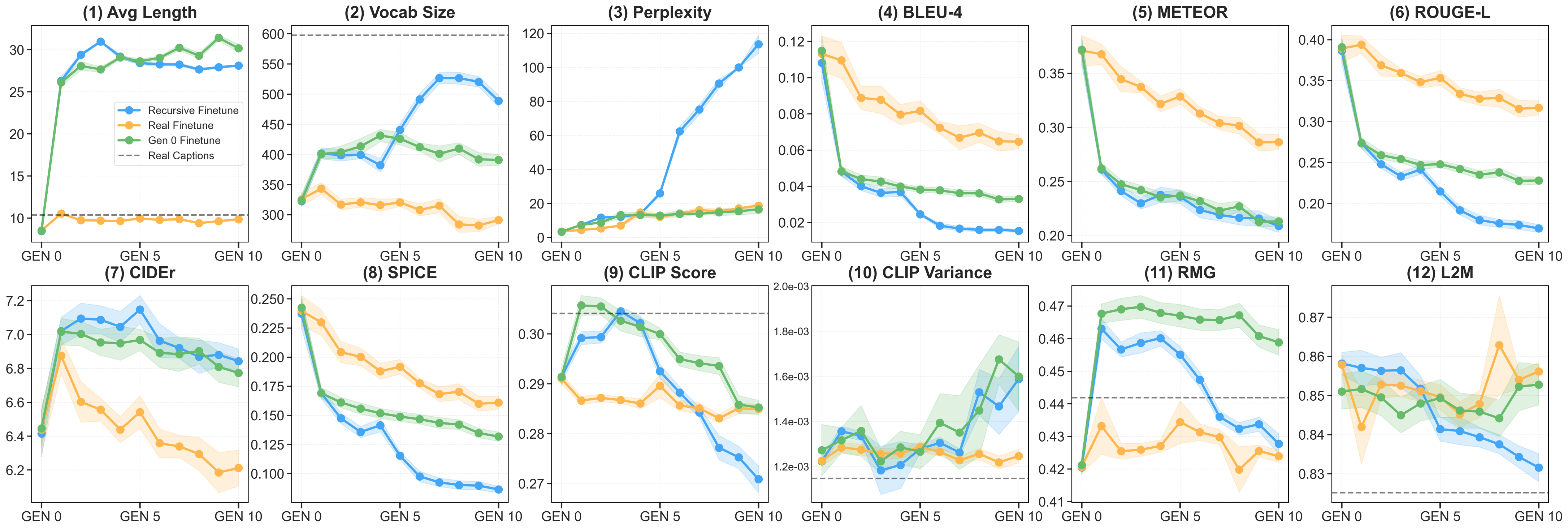}
    \caption{Recursive finetuning effect on the generated caption properties: Mean and standard deviation are calculated from 10 groups of 200 evaluation images randomly selected from the COCO evaluation set. Two baselines are shown: ``Gen 0 Finetune'' stands for finetuning on the synthetic data by the base (Generation 0) model. ``Real Finetune'' stands for finetuning on the MSCOCO data. The most prominent effect of recursive finetuning compared to the other two baselines is the vocabulary size and perplexity explosion (2, 3).}
    \Description{Graph showing how various caption quality metrics evolve during recursive finetuning of the vision-language model.}
    \label{fig:caption-metrics}
\end{figure*}

\begin{figure}[ht]
    \centering
    \includegraphics[width=\linewidth]{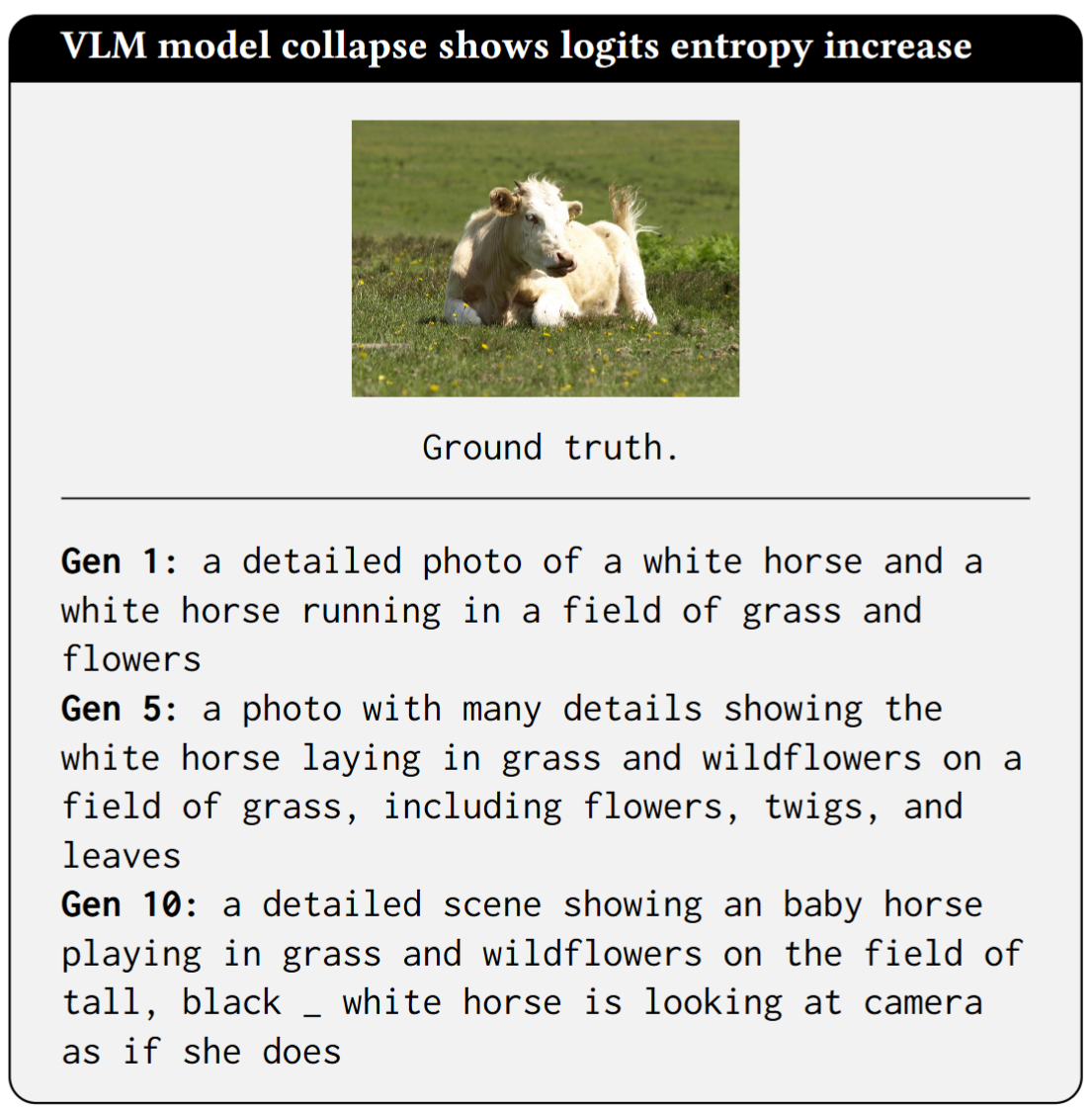}
    \caption{Generations from VLM recursive finetuning show increased language variance, details, but reduced grammatical coherence.}
     \Description{Graph showing VLM model collapse.}
    \label{fig:caption_exp}
\end{figure}

\textbf{Large performance discrepancy between synthetic training and MSCOCO data training at generation 1.} We observe a large metrics gap at generation 1 between synthetic and MSCOCO data training settings, where a full episode of synthetic data finetuning is finished in both recursive finetuning and ``Gen 0 Finetuning''. This gap is caused by the large discrepancies between the MSCOCO captions and the zero-shot (generation 0) synthetic captions. This gap is only seen in VLMs because the zero-shot image generated by the diffusion model is similar to the MSCOCO images, while the VLMs zero-shot caption is significantly longer than the MSCOCO captions. Thus, without the knowledge from the MSCOCO set captions, the synthetic finetunings quickly shift the distribution away from MSCOCO caption finetuning, and maintain such gaps for the remaining generations. \\\\
\textbf{Different collapsing behavior compared to LLMs.} Although sharing the same generative domain in language, VLMs in the image-captioning recursive finetuning show a different behavior compared to LLMs, as observed in previous research. For the vocabulary used for the test images, the recursive finetuning uses about 200 more tokens than ``Real Finetuning'' and 100 more than ``Gen 0 Finetuning'' at generation 10 (Figure ~\ref{fig:caption-metrics} subfigure 2). This observation suggests that the recursive finetuning causes the model to explore more tokens to explain the image, which is also reflected by the perplexity explosion in subfigure 3, indicating a more drastic softmax entropy increase. This observation is the opposite of the previously observed LLM collapse, where the vocabulary and perplexity used are significantly reduced \cite{modelcollapse, shumailov2023curse}. As for the captioning quality measure, we observe a more drastic drop of BLEU-4, ROUGE-L, METEOR, and SPICE scores in recursive finetuning, but a slower drop in terms of CIDEr which suggests that the synthetic data finetuning, including recursive finetuning, is able to maintain the semantics, but loses fluency or generates grammatically incorrect sentences. Since CIDEr is less sensitive to fluency, grammaticality, and details, which is the information lost in model collapse, the improved semantic alignment with the image can counteract and slow down the CIDEr drop.\\\\
\textbf{Variance increase.}
Different from the diffusion model collapse, we observe a steady increase in the variance of the generated captions (Figure \ref{fig:caption_exp}) due to the softmax entropy increase, reflected by the perplexity and vocabulary size increase. This observation also differs from the observation in single modality generative model collapse, where a variance collapse is a main property. The improved alignment is a shared property also seen in diffusion recursive finetuning. This phenomenon is due to the low sensitivity of CLIP embeddings towards grammar and more focus on object and property matching~\cite{Schrodi2024TwoEO}, which are improved through higher entropy and vocabulary usage.\\\\
\textbf{Alignment peaks then drops.}
In Figure ~\ref{fig:caption-metrics} subfigures 7 and 9, recursive finetuning setting shows an initial increase in CIDEr through generation 5 and a drop thereafter. Similar trend is also seen in the CLIP Score through generation 3. The initial increase of these metrics reflects a higher alignment between the two modalities, even than that of the real image caption pairs, due to boosted object alignment. The later slow decrease is a compounding effect of overfitting seen in the real finetuning (orange line), and the decrease in n-gram matching due to higher vocabulary size with similar average lengths.
Both synthetic finetuning cases boost the alignment above the real image and caption baseline, with the ``Gen 0 Finetuning'' having a peak at generation 1 and recursive finetuning at generation 3. These early generation peaks suggest that synthetic data training can be used to improve modality alignment when the recursive loop is shallow.

\subsection{Variance shift during recursive finetuning}
\label{subsec:variance}

Recent works \cite{modelcollapse, shumailov2023curse} have discovered that the variance decreases during LLMs' recursive finetuning. However, our experiments on multimodal models suggests that variance can be increased by the cross-modal interactions in VLM recursive finetuning.
\begin{figure}[ht]
    \centering
    \includegraphics[width=\linewidth]{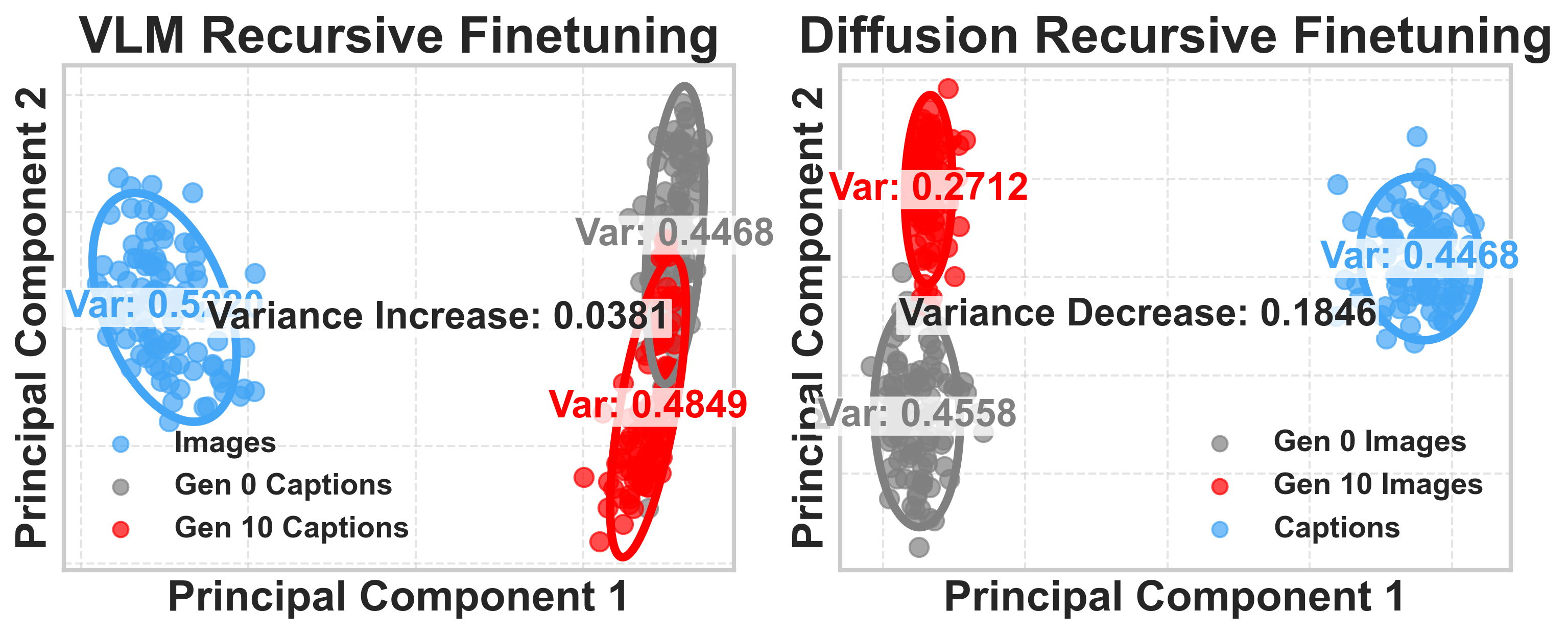}
    \caption{PCA of CLIP embeddings of VLM and diffusion recursive finetuning experiments. Variance changes in different directions for VLM and diffusion recursive finetuning.}
    \label{fig:PCA}
    \Description{PCA of CLIP embeddings of VLM and diffusion recursive finetuning experiments}
\end{figure}

 The different trend of variance shift in diffusion models and VLMs (Figure ~\ref{fig:PCA}) can be explained through the lens of information imbalance. As suggested in a recent work \cite{Schrodi2024TwoEO}, during contrastive VLM training, a model is trying to match the information content between vision and language. We claim that such an effect can also be extended to other vision-language generative paradigms, for example, the VLM and diffusion models. Both objectives explicitly or implicitly maximize the conditional likelihood from one modality to the other, and this process will bridge the variance gap between the two modalities. When the generation is from image to text, the high variance image will ``drag'' the variance of the text to a higher variance.
\subsection{Balanced bias-variance tradeoff mitigates model collapse}
\label{subsec:bias_variance}
\begin{figure}[ht]
    \centering
    \includegraphics[width=\linewidth]{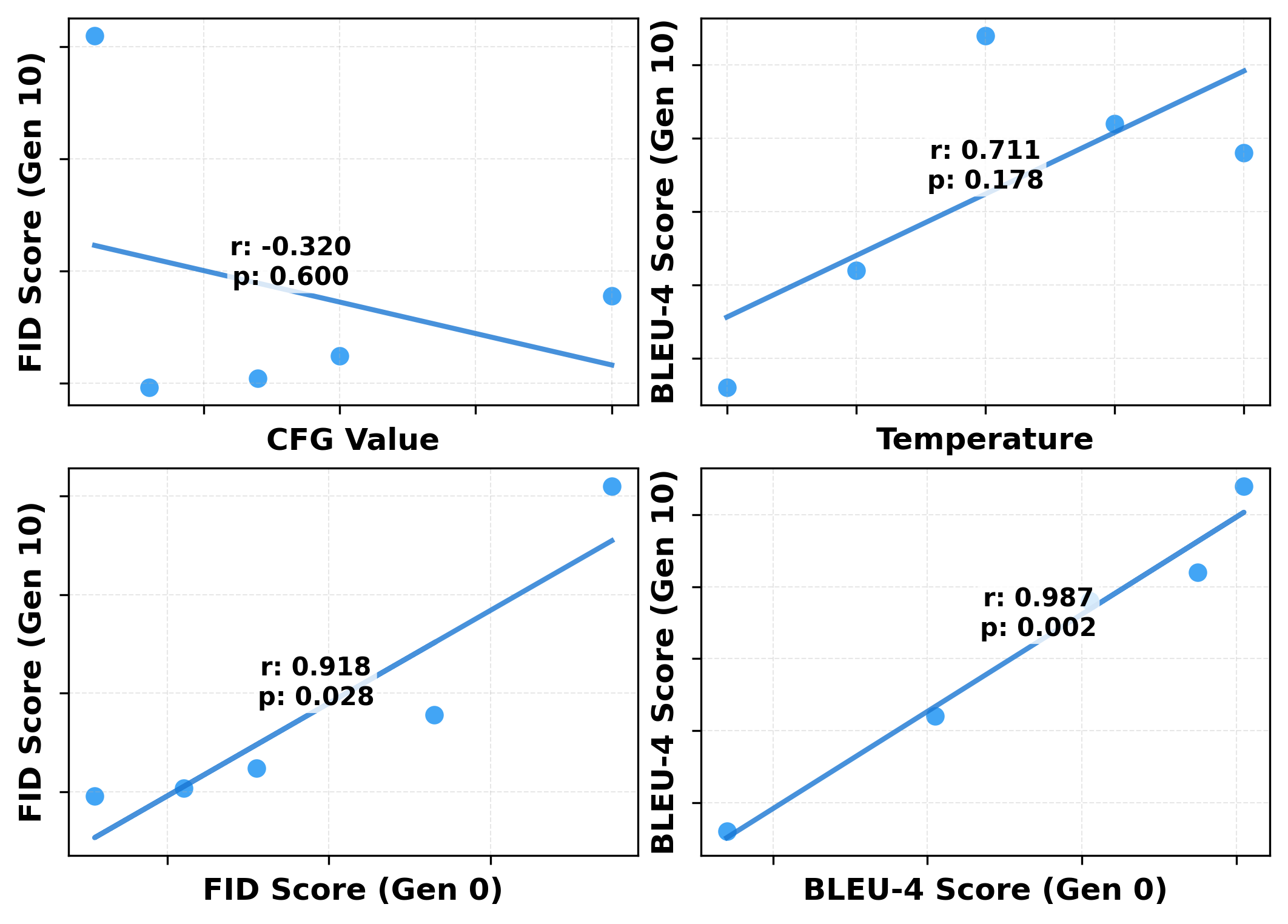}
    \caption{Correlation between bias-variance trade-off hyperparameters vs. Gen 10 quality measures. }
    \label{fig:pearson}
    \Description{Correlation between bias-variance trade-off hyperparameters vs. Gen 10 quality measures.}
\end{figure}
Both VLM and diffusion generative paradigms provide hyperparameters to control the bias-variance tradeoff during inference time without requiring additional training or affecting test-time complexity. This flexibility allows us to study how different trade-offs between bias and variance can affect the model collapse. We control the classifier-free guidance (CFG) scale \cite{ho2022classifier} and the temperature of the diffusion model and VLM during the decoding process, respectively. A higher CFG scale and lower temperature generate images and captions with higher bias and lower variance. In Figure~\ref{fig:pearson} we discover that, in general, the shifting bias and variance towards one direction (first row) does not correlate well with less model collapse, but the better generation quality in terms of FID and BLEU-4 score (peaks at a balanced bias-variance trade-off) in Gen 0 has a high correlation with less model collapse. These correlations indicate that simple quality measures such as FID for images and BLEU-4 for captions can be good indicators for synthetic data robustness against model collapse, and the optimal values for both metrics require a balanced bias-variance trade-off.

\subsection{More decoding budgets mitigate model collapse}
\label{subsec:decoding_budgets}
\begin{figure}[ht]
\centering
\includegraphics[width=\linewidth]{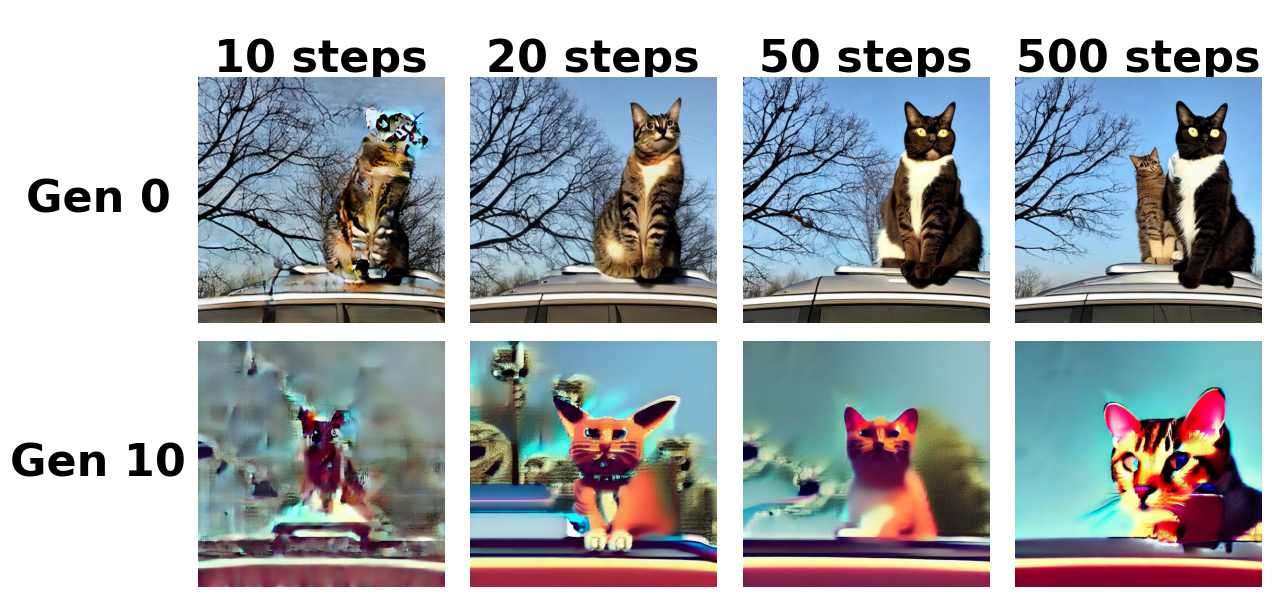}
\caption{Increasing denoising budget for diffusion generation reduces model collapse, as seen in text-to-image generations for "a cat is sitting on top of a vehicle." The best quality (FID) of the generated image is the one with 50 denoising steps.}
\Description{Graph showing how increasing the number of generation steps improves the stability of generated images and reduces model collapse.}
\label{fig:cats}
\end{figure}

Decoding budgets have been a recent focus on LLMs' inference time scaling~\cite{s1, deepseekr1, gpt4o}. This trend drives us to study the correlation between the decoding budgets and the model's robustness towards model collapse. We first notice that generated samples have the best FIDs at 50 denoising steps for diffusion models, while the FID remains stable for more denoising steps. However, at generation 10 of recursive finetuning, the image quality consistently increases as we increase the number of generation steps (Figure ~\ref{fig:cats}). 

\subsection{Model and hyperparameter diversity mitigates model collapse}
\label{subsec:model_diversity}
In our model diversity experiments, we observed that incorporating a variety of models with different hyperparameters and architectures led to a more robust recursive training process. Both approaches mitigates the model collapse in VLMs and diffusion recursive finetuning, as shown in Table~\ref{tab:diversity}.
Note that this diversity study does not mix VLMs with diffusion models, which is studied as relabeling and joint finetuning in the following sections.
\begin{table}[ht]
\footnotesize
\caption{Effect of model diversity on quality metrics at generation 10. Both hyperparameter diversity and architectural diversity mitigate model collapse.}
\label{tab:diversity}
\centering
\begin{tabular}{lccc}
\toprule
\textbf{Settings} & \textbf{Gen 10 FID↓} & \textbf{Gen 10 BLEU@4↑}\\
\midrule
\multicolumn{3}{l}{\textbf{Diffusion Model Recursive Finetuning}}\\
\midrule
SD v1.4 & 253.2 ± 5.4 & -\\
\midrule
\multicolumn{3}{l}{\textit{Hyperparameter Diversity}}\\
\midrule
SD v1.4 (2 CFGs, 3, 7) & 241.3 ± 6.2 & -\\
\midrule
\multicolumn{3}{l}{\textit{Architectural Diversity}}\\
\midrule
SD v1.4 ``Gen 0 Finetune'' & 242.1 ± 5.8 & -\\
SD v1.4 + SD v1.5 & 207.3 ± 7.5 & -\\
SD v1.4 + Flux.1 & \textbf{194.2 } & -\\
\midrule
\multicolumn{3}{l}{\textbf{VLM Model Recursive Finetuning}}\\
\midrule
BLIP-2 & - & 0.017 ± 0.012\\
\midrule
\multicolumn{3}{l}{\textit{Hyperparameter Diversity}}\\
\midrule
BLIP-2 (2 temperatures, 1, 0.7) & - & 0.024 ± 0.014\\
\midrule
\multicolumn{3}{l}{\textit{Architectural Diversity}}\\
\midrule
BLIP-2 ``Gen 0 Finetune'' & - & 0.039 ± 0.0012\\
BLIP-2 + MiniCPM v2.6 & - & \textbf{0.071}\\
\bottomrule
\end{tabular}
\end{table}

\subsection{Relabeling mitigates model collapse}
\label{subsec:relabeling}
\begin{table}[ht]
\small
\caption{Quality metrics at generation 10 after recursive fine-tuning, demonstrating relabeling reduces model collapse in both diffusion and VLM models.}
\label{tab:relabeling}
\centering
\begin{tabular}{lccc}
\toprule
\textbf{Settings} & \textbf{Gen 10 FID↓} & \textbf{Gen 10 BLEU@4↑}\\
\midrule
\multicolumn{3}{l}{\textbf{Diffusion Model Recursive Fine-tuning}}\\
\midrule
Pure Recursive & 253.2 ± 5.4 & -\\
VLM Relabeling & \textbf{223.1 ± 8.2} & -\\
\midrule
\multicolumn{3}{l}{\textit{With Prefix Tags}}\\
\midrule
"Synthetic image of" & 244.5 ± 4.3 & -\\
"\#"  & 245.2 ± 4.8 & -\\
"Synthetic N image of" & 239.8 ± 11.0 & -\\
"A bright image of"  & \textbf{238.9 ± 6.5} & -\\
\midrule
\multicolumn{3}{l}{\textbf{VLM Model Recursive Fine-tuning}}\\
\midrule
Pure Recursive & - & 0.017 ± 0.012\\
Diffusion Relabeling & - & \textbf{0.032 ± 0.018}\\
\bottomrule
\end{tabular}
\end{table}
\textbf{Relabeling.}
By introducing VLM models into the process of diffusion recursive training and vice versa, we made the conditional modality dynamic by relabeling it after every generation. For example, in the diffusion recursive training, a frozen VLM updates the caption of the newly generated synthetic image. This approach mitigates model collapse, shown as improved quality after 10 generations of recursive training (Table ~\ref{tab:relabeling}).\\\\ 
\textbf{Prefix tag.}
In addition to relabeling with generative models, we provide two simple baselines that also change the conditioning caption in diffusion pretraining by rules. We set up 2 approaches: 1. We simply append the prefix word "Synthetic" to all gen 1 to gen 10 captions, indicating the generated images are synthetic. The intuition behind the setting is that we assume the diffusion model already has some pretrained knowledge on the word ``synthetic image of'', and we provide better alignment by adding this information. 2. We append "Synthetic 1", "Synthetic 2" to all gen 1 to gen 10 captions to provide additional information about the synthetic generations. In addition, we also conduct ablations on using other placeholder tokens such as ``\#'' or semantic tokens such as ``bright.'' Similarly, in VLM recursive training, the frozen diffusion model generates images from the newly generated synthetic captions. Since the baseline and ablation experiment designs are not as intuitive, we do not provide them for VLM training.
\textbf{Theoretical basis for relabeling effectiveness.} The effectiveness of relabeling stems from providing a novel grounding for the synthetic data, distinct from the model's prior association with the conditioning texts or images. Instead of introducing a contradicting "one-to-many" mapping from the conditions to the synthetic data recursively, relabeling introduces a fresh condition. This can alleviate the competition inherent in forcing synthetic data into pre-established real-data relationships. The success of both relabeling and even simpler prefix tagging supports this idea, as both introduce a new conditioning signal. However, relabeling with a generative model offers a grounding that is semantically closer to the synthetic data, thereby guiding the model's distribution shift in a less drastic manner compared to mere tagging.

\subsection{Joint finetuning exacerbates model collapse}
\label{subsec:joint_training}

In our Setting 3 experiments (Figure \ref{fig:settings} subfigure 3), we jointly fine-tune both the diffusion model and VLM on synthetic data pairs. This approach differs from the relabeling method (Setting 2) in that both models evolve together rather than having one model remain frozen. Surprisingly, our findings indicate that joint fine-tuning actually worsens model collapse compared to the relabeling approach.

When both models are allowed to adapt to each other's outputs simultaneously, we observe an accelerated co-degradation process. Without a stable reference point (frozen model), both models reinforce each other's biases and errors, leading to more rapid distributional drift. This mutual adaptation creates a feedback loop that amplifies rather than mitigates the collapse process.

Table \ref{tab:joint} summarizes the results of joint fine-tuning compared to both single-model recursive fine-tuning and relabeling approaches. The joint approach shows significant degradation in generation quality across iterations, with faster deterioration of FID scores and lower BLEU@4 scores at generation 10 compared to the relabeling method.

\begin{table}[ht]
\small
\caption{Comparison of generation quality at generation 10 across different fine-tuning approaches. Joint fine-tuning of both models shows accelerated degradation in quality metrics compared to relabeling with a frozen model.}
\label{tab:joint}
\centering
\begin{tabular}{lccc}
\toprule
\textbf{Approach} & \textbf{Gen 10 FID↓} & \textbf{Gen 10 BLEU@4↑}\\
\midrule
Single Model Recursive & 253.2 & 0.017\\
With Relabeling & \textbf{223.1} & \textbf{0.032}\\
Joint Fine-tuning & 312.5 & 0.009\\
\bottomrule
\end{tabular}
\end{table}

The joint fine-tuning results demonstrate that maintaining at least one frozen model is crucial when using synthetic data in multi-modal contexts. The frozen model serves as an anchor that prevents both models from drifting too far from the original data distribution. This finding highlights the importance of stability in multi-modal generative systems and suggests that completely self-contained recursive learning loops without external anchoring can rapidly destabilize.

\section{Conclusion}
This study investigates multimodal synthetic data finetuning and multi-modal model collapse. Our analysis provides insights into understanding and mitigating model collapse in recursive fine-tuning. We highlight several key findings: (1) variance does not always collapse during model collapse, as evidenced in VLM image captioning (Section~\ref{subsec:variance}); (2) increased decoding budgets generate more robust synthetic datasets (Section~\ref{subsec:decoding_budgets}); (3) commonly used quality measures correlate well with robustness against model collapse (Section~\ref{subsec:bias_variance}); and (4) model diversity is crucial for maintaining stability in recursive fine-tuning (Section~\ref{subsec:model_diversity}).

In addition, our experiments demonstrate that relabeling with a frozen model significantly slows model collapse compared to single-model recursive fine-tuning, while joint fine-tuning without a frozen model accelerates model collapse. This finding highlights the critical importance of maintaining human-authored data-grounded, frozen models in an autonomous multi-agent self-training loop to mitigate model collapse.

\section{Safe and Responsible Innovation Statement}
Our research investigates the risks of model collapse in multimodal vision-language generative systems when trained on synthetic data. While synthetic data offers benefits, its uncontrolled recursive use can lead to performance degradation. This work addresses potential societal impacts by exploring methods to mitigate this collapse, ensuring the reliability of future AI systems that may rely on self-generated data. Ethical considerations include the potential for synthetic data to perpetuate or amplify biases present in the initial models or training data. Our findings on the benefits of model diversity and frozen grounding models contribute to responsible deployment by suggesting architectural strategies to maintain stability and prevent uncontrolled drift in autonomous multi-agent AI environments. By providing guidelines for robust synthetic data training, we aim to encourage responsible innovation in multimodal interaction research, minimizing the risks of misuse or unintended negative consequences arising from model collapse.
\bibliographystyle{ACM-Reference-Format}
\bibliography{bibliography}

\clearpage
\appendix

\section{Evaluation Metrics}
\label{appendix:metrics}

This appendix details the evaluation metrics used to assess the performance of our image generation models. We employed a combination of quantitative metrics to evaluate various aspects of the generated images, including image quality, diversity, and alignment with textual descriptions.

\subsection{Contrast}

The contrast of the generated images was measured using the Root Mean Square (RMS) contrast, defined as:

$$
\text{Contrast} = \frac{\sigma_{gray}}{\mu_{gray}}
$$

where $\sigma_{gray}$ is the standard deviation of the grayscale image, and $\mu_{gray}$ is its mean.

\subsection{Brightness}

The brightness of the generated images was calculated as the average pixel intensity of the grayscale image, scaled to the range $[0, 255]$:

$$
\text{Brightness} = \mu_{gray} \times 255
$$

\subsection{Color Distribution}

The color distribution of the generated images was analyzed by calculating the average RGB values, color saturation, and color standard deviation. The average RGB values were computed as:

$$
\text{Avg R} = \mu_{R}
$$

$$
\text{Avg G} = \mu_{G}
$$

$$
\text{Avg B} = \mu_{B}
$$

where $\mu_{R}$, $\mu_{G}$, and $\mu_{B}$ are the means of the red, green, and blue color channels, respectively. The color saturation was calculated as the mean difference between the maximum and minimum channel values:

$$
\text{Saturation} = \mu(\max(R, G, B) - \min(R, G, B))
$$

The color standard deviation was computed as the average standard deviation of the RGB channels:

$$
\text{Color Std} = \frac{\sigma_{R} + \sigma_{G} + \sigma_{B}}{3}
$$

where $\sigma_{R}$, $\sigma_{G}$, and $\sigma_{B}$ are the standard deviations of the red, green, and blue color channels, respectively.

\subsection{Fréchet Inception Distance (FID)}

The Fréchet Inception Distance (FID) was used to measure the similarity between the distribution of generated images and the distribution of real images. FID is defined as:

$$
\text{FID} = ||\mu_x - \mu_g||^2 + \text{Tr}(\Sigma_x + \Sigma_g - 2(\Sigma_x \Sigma_g)^{1/2})
$$

where $\mu_x$ and $\mu_g$ are the mean feature vectors of the real and generated images, respectively, and $\Sigma_x$ and $\Sigma_g$ are their covariance matrices.

\subsection{Inception Score (IS)}

The Inception Score (IS) was used to evaluate the quality and diversity of the generated images. IS is defined as:

$$
\text{IS} = \exp(E_{x \sim p_g} [D_{KL}(p(y|x) || p(y))])
$$

where $p_g$ is the distribution of generated images, $p(y|x)$ is the conditional class distribution given an image $x$, and $p(y)$ is the marginal class distribution.

\subsection{CLIP Score}

The CLIP score was used to measure the alignment between the generated images and their textual descriptions. The CLIP score is defined as the cosine similarity between the CLIP embeddings of the generated images and their corresponding captions. The implementation is identical to the original paper~\cite{hessel_clipscore_2022}.

\subsection{Relative Modality Gap (RMG)}

The Relative Modality Gap (RMG) measures the relative dissimilarity between image and text embeddings. The implementation is identical to the original paper~\cite{Schrodi2024TwoEO}.

\subsection{L2 Distance Modality Gap (L2M )}

L2M is the L2 distance between the mean of the image and text embedding features. The implementation is identical to the original paper~\cite{liang2022mind}.

\subsection{CLIP Variance}

CLIP Variance measures the variance of CLIP image embeddings.

\section{Detailed Experimental Setup}
\label{appendix:experimental_details}

This section provides comprehensive details about our experimental setup, including data preprocessing, hyperparameter configurations, and model specifications.

\subsection{Data Preprocessing}
\label{appendix:data_preprocessing}

For our experiments, we used a subset of the MSCOCO training set, which contains 1000 image-caption pairs. We preprocessed the dataset as follows:

\begin{enumerate}
    \item Images were resized to 256×256 pixels for input to the diffusion models
    \item We removed instances that are not gender-neutral for gender bias analysis
\end{enumerate}

Data augmentations were minimized to ensure that any observed effects were due to the recursive fine-tuning process rather than data preprocessing artifacts.

\subsection{Hyperparameter Settings}
\label{appendix:hyperparameters}

The following hyperparameters were used in our experiments:

\textbf{Stable Diffusion 1.4:}
\begin{itemize}
    \item Learning rate: 1e-5
    \item Batch size: 8
    \item Training steps per generation: 2000
    \item CFG scale default: 7.0 (unless specified otherwise in experiments)
    \item Diffusion steps: 50 (unless specified otherwise in experiments)
\end{itemize}

\textbf{BLIP-2:}
\begin{itemize}
    \item Learning rate: 5e-6
    \item Batch size: 16
    \item Training steps per generation: 2000
    \item Beam search size: 4 (unless specified otherwise in experiments)
    \item Temperature: 1.0 (unless specified otherwise in experiments)
\end{itemize}

\subsection{Model Specifications}
\label{appendix:models}

\textbf{Diffusion Models:}
We primarily used Stable Diffusion 1.5, which is a latent diffusion model trained on billions of image-text pairs. For architectural diversity experiments, we also included:
\begin{itemize}
    \item Stable Diffusion 1.4: An earlier version with 860M parameters
    \item Flux.1: A newer diffusion model with improved handling of certain visual properties
\end{itemize}

\textbf{Vision-Language Models:}
Our main VLM was BLIP-2, which uses a frozen image encoder and a large language model for captioning. For diversity experiments, we also used:
\begin{itemize}
    \item BLIP: The predecessor to BLIP-2 with a different architecture
    \item MiniCPM-Llama3-V 2.5: A smaller but efficient multi-modal model
\end{itemize}

All models were implemented using PyTorch and trained and run on a NVIDIA A100 GPU with 40GB of memory and a NVIDIA RTX 4090 GPU.

\end{document}